\documentclass{article}

\usepackage{microtype}
\usepackage{graphicx}
\usepackage{subcaption}
\usepackage{enumitem}
\usepackage{booktabs} %
\usepackage[most]{tcolorbox}
\usepackage{hyperref}

\usepackage{xcolor}

\usepackage[preprint]{icml2026}

\usepackage{amsmath}
\usepackage{amssymb}
\usepackage{mathtools}
\usepackage{amsthm}

\usepackage{balance}

\usepackage[capitalize,noabbrev]{cleveref}

\theoremstyle{plain}

\theoremstyle{definition}

\theoremstyle{remark}

\usepackage[textsize=tiny]{todonotes}

\icmltitlerunning{Not All Layers Need Tuning: Selective Layer Restoration Recovers Diversity}

\begin{document}

\twocolumn[
  \icmltitle{Not All Layers Need Tuning: Selective Layer Restoration Recovers Diversity}

  \icmlsetsymbol{equal}{*}

  \begin{icmlauthorlist}
    \icmlauthor{Bowen Zhang}{soc}
    \icmlauthor{Meiyi Wang}{soc}
    \icmlauthor{Harold Soh}{soc,ssi}
  \end{icmlauthorlist}

  \icmlaffiliation{soc}{Department of Computer Science, National University of Singapore, Singapore}
  \icmlaffiliation{ssi}{Smart Systems Institute, National University of Singapore, Singapore, Singapore}

  \icmlcorrespondingauthor{Bowen Zhang}{bowenzhang@comp.nus.edu.sg}
  \icmlkeywords{Machine Learning, ICML, Natural Language Processing, Large Language Models, Text Generation, Deep Learning, Artificial Intelligence, Training-free, Generation Diversity}

  \vskip 0.3in
]

\printAffiliationsAndNotice{}  %

\begin{abstract}
Post-training improves instruction-following and helpfulness of large language models (LLMs) but often reduces generation diversity, which leads to repetitive outputs in open-ended settings, a phenomenon known as mode collapse. Motivated by evidence that LLM layers play distinct functional roles, we hypothesize that mode collapse can be localized to specific layers and that restoring a carefully chosen range of layers to their pre-trained weights can recover diversity while maintaining high output quality. To validate this hypothesis and decide which layers to restore, we design a proxy task---Constrained Random Character (CRC)---with an explicit validity set and a natural diversity objective. Results on CRC reveal a clear diversity–validity trade-off across restoration ranges and identify configurations that increase diversity with minimal quality loss. Based on these findings, we propose Selective Layer Restoration (SLR), a training-free method that restores selected layers in a post-trained model to their pre-trained weights, yielding a hybrid model with the same architecture and parameter count, incurring no additional inference cost. Across three different tasks (creative writing, open-ended question answering, and multi-step reasoning) and three different model families (Llama, Qwen, and Gemma), we find SLR can consistently and substantially improve output diversity while maintaining high output quality.
\end{abstract}

\begin{figure}[t!]
    \centering
    \includegraphics[width=0.48\textwidth]{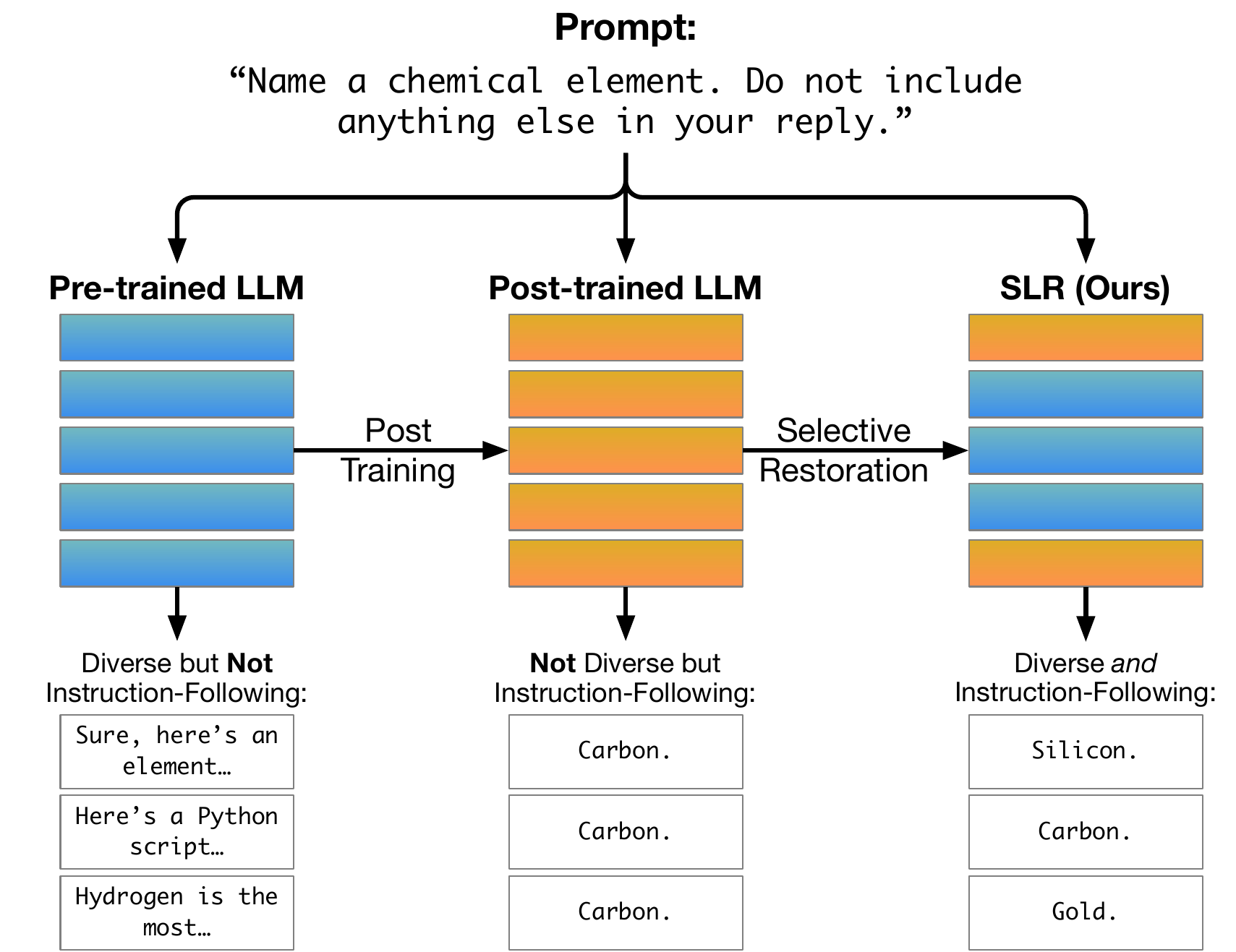}
    \caption{Pre-trained LLMs are diverse but have weak instruction-following ability, leading to low-quality responses. Post-trained LLMs show strong instruction adherence and high output quality but suffer from mode collapse. Selective layer restoration (SLR) restores the diverse modes existing in pre-trained LLMs while maintaining high output quality.}
    \label{fig:intro_fig}
\end{figure}

\section{Introduction}
Post-training, including instruction tuning~\cite{wei2021finetuned, sanh2021multitask} and RLHF~\cite{ouyang2022rlhf}, is a double-edged sword. It improves output quality (e.g., instruction adherence and helpfulness) in large language models (LLMs), but is also observed to induce mode collapse~\cite{kirk2023understanding,o2024attributing}; the post-trained model assigns disproportionately high probability mass to a narrow range of responses (the ``mode'') among many valid alternatives. This leads to a \textit{reduced output diversity} and is undesirable in settings where multiple distinct responses are valued (e.g., creative writing, open-ended question answering, and reasoning), thereby limiting their effectiveness in these applications.

Existing approaches to mitigating mode collapse can be grouped into three broad families: \textit{decoding-based}~\cite{nguyen2024turning, basu2020mirostat,tang2024top}, \textit{prompt-based}~\cite{zhang2025verbalized, ge2024scaling, summers2023brainstorm}, and \textit{training-based} methods~\cite{huang2025diversity_aware_policy_opt,li2024preserving_diversity_sft}. Each comes with distinct limitations;  decoding-based methods cannot recover diversity lost during post-training, prompt-based methods increase inference cost, and training-based methods require expensive retraining. In this work, we seek a method that (i) recovers diversity effectively, (ii) is computationally efficient during adaptation/training, and (iii) introduces no additional inference-time cost.

We draw on two empirical observations. First, LLM layers exhibit different functional roles~\cite{meng2022rome, tenney2019bert_pipeline, song2025demystifying} (e.g., factual associations tend to be localized to mid-depth layers~\cite{meng2022rome}). Second, much of the model’s knowledge is acquired during pretraining, while post-training primarily steers which pretrained behaviors are expressed and can suppress others (catastrophic forgetting)~\cite{kotha2024forgetting, li2024revisiting}. Based on these findings, we hypothesize that the mode collapse induced by post-training may be localized to specific layers, and that \textbf{selectively restoring} a carefully chosen set of layers to their \textbf{pre-trained parameters} can recover output~\textit{\textbf{diversity}} while preserving high output~\textit{\textbf{quality}}.

To test our hypothesis and make it actionable, we reduce it to a concrete selection problem: \textit{which layers should be restored?} We cast this as a \textit{diversity–quality} constrained optimization problem, where the objective is to maximize output diversity while maintaining output quality above a minimum threshold. 

This setup entails solving two challenges. As posed, the general problem involves searching over arbitrary subsets of layers, which is a large combinatorial search-space. We make the problem tractable by restoring a \textit{contiguous interval} of layers. A second issue is that directly estimating diversity and quality for each candidate restoration interval on complex downstream tasks (such as creative writing) would be expensive and difficult, especially for ``quality'' which is often noisy or subjective. To address this, we introduce a simple proxy task, \textbf{CRC} (\textbf{C}onstrained \textbf{R}andom \textbf{C}haracter): random digit and letter generation under strict output constraints (e.g., ``Generate a random number in the range [0, 5]''). On CRC, both diversity and quality are naturally and unambiguously defined, enabling efficient exploration of restoration intervals and guiding the choice of layers to restore. The proxy serves two purposes---it provides evidence that selective restoration can recover diversity with limited quality loss, thus supporting our layer-localization hypothesis, and it guides the choice of layers to restore for downstream tasks.

Guided by this proxy, we propose \textbf{Selective Layer Restoration (SLR)}, which constructs a hybrid model by restoring a chosen interval of layers in a post-trained LLM to their pre-trained parameters. SLR directly meets our desiderata since it is \textbf{training-free} (operating directly on existing checkpoints), \textbf{computationally efficient}, and \textbf{keeps inference cost unchanged} (it preserves the architecture and parameter count). We evaluate SLR on three downstream tasks that require both diversity and quality---creative writing, open-ended question answering, and multi-step reasoning---and across three model families (Qwen~\cite{team2024qwen2}, Gemma~\cite{team2024gemma}, and Llama~\cite{grattafiori2024llama}). Across these settings, we find consistent and substantial output diversity gains while maintaining high output quality. Moreover, we further show that SLR is complementary to decoding- and prompt-based methods, yielding additional gains when combined. Finally, we conduct ablation studies to confirm the importance of CRC-guided restoration interval selection.

In summary, our paper makes the following contributions:
\begin{itemize}[noitemsep]
    \item SLR, a simple yet effective training-free weight-space intervention that restores an interval of layers in a post-trained LLM to their pre-trained parameters, without changing architecture or additional inference cost. It is also complementary to common decoding- and prompt-based diversity interventions.
    \item CRC, a proxy task with an explicit validity set and natural diversity objective. CRC provides evidence for our layer-localization hypothesis and guides the selection of restoration intervals.
    \item Empirical evidence that demonstrates the effectiveness of SLR across three downstream tasks and three model families.
\end{itemize}

More broadly, our findings support a modular view of post-trained LLMs. In particular, post-training does not affect all layers equally and the behaviors associated with mode collapse appear to be localized to specific layers. Our work shows that desirable properties of pre-trained models like diversity can be recovered by targeted interventions.

\section{Related Work}
In this section, we summarize related work on mode collapse in post-trained LLMs, existing mitigation strategies, layer-wise functional roles in transformer models, and model merging techniques. These lines of research provide the context and motivation for our approach.

\paragraph{Mode Collapse in Post-trained LLMs} It is widely observed that post-trained LLMs suffer from mode collapse, have significantly lower output diversity compared to their pre-trained counterparts~\cite{kirk2023understanding,o2024attributing, xiao2025algorithmic}. Mode collapse is often attributed to factors such as KL-regularized optimization used in RLHF~\cite{xiao2025algorithmic} and bias in the post-training data~\cite{zhang2025verbalized}. Broadly, post-training can be viewed as steering the pretrained distribution toward responses favored by supervision or preference data, thereby concentrating probability mass on a subset of already-plausible modes and suppressing others~\cite{kotha2024forgetting, song2025demystifying}.

\paragraph{Mitigating Mode Collapse.} Existing methods to mitigate mode collapse can be categorized into three families: (i) Decoding-based methods work by increasing stochasticity (increasing temperature~\cite{renze2024effect}), truncating the distribution by different cutoff thresholds~\cite{nguyen2024turning, tang2024top}, or using heuristics like controlling output perplexity~\cite{basu2020mirostat}. Since they do not directly alter the model parameters, they are limited in recovering behaviors that may be suppressed during post-training. (ii) Prompt-based methods rely on hand-crafted prompt templates to directly modify the input, e.g., explicitly requesting multiple distinct answers~\cite{zhang2025verbalized}, enforcing different perspectives~\cite{ge2024scaling}. However, their effectiveness is sensitive to the prompt wording and task context, and the increased prompt length will increase the inference cost. (iii) Training-based methods modify post-training methods to explicitly encourage diversity, e.g., via objective design~\cite{huang2025diversity_aware_policy_opt,li2024preserving_diversity_sft}, data collection strategies~\cite{song2024scaling}. Such approaches require re-training and careful tuning, often with non-trivial computational and data cost. In contrast, SLR operates directly in weight space on existing checkpoints, improving diversity without additional training or inference-time overhead.

\paragraph{Functional Roles of LLM layers.} Prior works~\cite{tenney2019bert_pipeline, meng2022rome,song2025demystifying} suggest that transformer-based LLMs exhibit depth-dependent functional specialization. It is found that earlier layers correlate more with surface/lexical and local syntactic features, and deeper layers more with higher-level semantics~\cite{tenney2019bert_pipeline, song2025demystifying}. Complementing this, model editing work provides more causal evidence of localization: e.g., ROME and related analyses~\cite{meng2022rome} identify that knowledge storage and retrieval can be localized to mid-depth layers, and modifying them can effectively edit the knowledge stored in the LLMs. Together, these findings motivate our hypothesis that mode collapse may also be localized to specific LLM layers.

\paragraph{Model Merging.} Model merging combines checkpoints in weight space without changing the architecture or increasing per-token inference cost. A common approach is (optionally weighted) parameter averaging across checkpoints~\cite{wortsman2022model, yu2024language, davari2024model}, and related work studies other merging operators that combine models at a finer granularity, such as editing or stitching subsets of layers or sub-layer components to transfer or compose capabilities~\cite{wortsman2022model, yu2024language, davari2024model, hu2025navigating,bandarkar2024layer}. Although much of this literature focuses on combining multiple post-trained or task-specialized checkpoints (``experts'')~\cite{wortsman2022model, yu2024language, davari2024model}, some recent studies explore merging a post-trained model with its pre-trained counterpart and observe trade-offs between quality and diversity~\cite{hu2025navigating}. In our work, SLR performs structured restoration between a post-trained model and its pre-trained counterpart with the explicit goal of improving the diversity–quality trade-off, and uses CRC to guide which components to merge.

\section{Methodology}
\label{sec:methodology}

This section presents our primary contribution, SLR, a training-free method for mitigating mode collapse in post-trained LLMs by selectively restoring a contiguous interval of transformer layers to their pre-trained weights. 

\subsection{SLR: Selective Layer Restoration}
Let $M_{\text{pre}}$ denote a pre-trained language model and $M_{\text{post}}$ its post-trained counterpart, sharing the same transformer architecture with $N$ layers of transformer blocks. We write their layer stacks as
\begin{align*}
M_{\text{pre}} & = \{L^{\text{pre}}_0,\dots,L^{\text{pre}}_{N-1}\}\\
M_{\text{post}} & =\{L^{\text{post}}_0,\dots,L^{\text{post}}_{N-1}\}.
\end{align*}
For any layer interval $0 \le i \le j \le N-1$, define the hybrid model $M_{i:j}$ by replacing the corresponding post-trained layers with pre-trained layers:
\begin{align*}
M_{i:j} & \triangleq R_{i:j}(M_{\text{post}}, M_{\text{pre}}),\\
L^{i:j}_k & =
\begin{cases}
L^{\text{pre}}_k, & i \le k \le j,\\
L^{\text{post}}_k, & \text{otherwise}.
\end{cases}
\end{align*}
Given a task $\mathcal{T}$ (e.g., creative writing), we associate each model $M$ with a \emph{quality} score $Q_{\mathcal{T}}(M)$ and a \emph{diversity} score $D_{\mathcal{T}}(M)$. \textbf{S}elective \textbf{L}ayer \textbf{R}estoration (SLR) therefore works by identifying the restoration interval $(i,j)$ that maximizes diversity while maintaining quality above a threshold $q_\mathrm{min}$ and constructs the hybrid model accordingly:
\[
\max_{0 \le i \le j \le N-1} \; D_{\mathcal{T}}(M_{i:j})
\quad
\text{s.t.}
\quad
Q_{\mathcal{T}}(M_{i:j}) \ge q_\mathrm{min}.
\]

\subsection{CRC: Constrained Random Character}
Directly solving the above problem on downstream tasks, such as creative writing, is challenging: evaluating $Q$ and $D$ requires multiple generations per prompt and expensive task-specific assessment. Instead, we design a simple proxy task,  \textbf{C}onstrained \textbf{R}andom \textbf{C}haracter (\textbf{CRC}).

CRC is designed so that both diversity and quality are \emph{unambiguous, discrete, and automatically measurable}.
Each prompt specifies a finite valid output set and requires \emph{a single-token answer} from that set. In this way, CRC avoids expensive sampling-based evaluation by directly using the model's predicted distribution at the first generation step.
We instantiate two families of constraints:
\begin{itemize}[noitemsep]
\item \textbf{Digit constraints}, e.g., ``\textsf{\small Generate a random integer in the range [0, 5]. Do not include anything else in your reply.}'', with valid set $\{0,1,2,3,4,5\}$ and 
\item \textbf{Letter constraints}, e.g., ``\textsf{\small Generate a random letter from A to G (inclusive). Do not include anything else in your reply.}'', with valid set $\{A,B,C,D,E,F,G\}$.
\end{itemize}
This construction removes semantic ambiguity. For each prompt, the set of valid outputs is explicitly known, and the valid set can be varied, enabling us to generate numerous prompts with different valid sets while using the same evaluation method. In practice, we construct 20 prompts for each constraint family by sampling over valid ranges, e.g., [0,5], [1,7].

Let $D$ denote the set of prompts. For a prompt $x\in D$ with a valid token set $\mathcal{V}(x)$, let $p_M(\cdot \mid x)$ be the model's next-token distribution at the \emph{first generated position} (i.e., immediately after the input).
We define the \emph{quality} score as the probability mass assigned to valid outputs,
\[
Q_x(M) \triangleq \sum_{v\in \mathcal{V}(x)} p_M(v\mid x)
\]
and the average CRC quality score as, 
\[
Q_{\mathrm{CRC}}(M) \triangleq \frac{1}{|D|}\sum_{x\in D} Q_x(M),
\]
This measures strict instruction adherence under the constraint. Conditioned on producing a valid token, the induced distribution over valid outputs is
\[
\tilde{p}_M(v \mid x) \triangleq \frac{p_M(v \mid x)}{\sum_{u \in \mathcal{V}(x)} p_M(u \mid x)} \quad \text{for } v \in \mathcal{V}(x),
\]
and we define the \emph{diversity} score as the entropy over valid outputs
\[
D_x(M) \triangleq - \sum_{v\in \mathcal{V}(x)} \tilde{p}_M(v\mid x)\log \tilde{p}_M(v\mid x),
\] 
and the average CRC diversity score,
\[
D_{\mathrm{CRC}}(M) \triangleq \frac{1}{|\mathrm{CRC}|}\sum_{x\in \mathrm{CRC}} D_x(M).
\]

Applying CRC using a na\"ive grid search over all possible interval configurations, i.e., $\frac{N(N-1)}{2}$ intervals given $N$ layers, takes around 30 minutes, 45 minutes, and 1 hour for Llama, Qwen, and Gemma, respectively, on one single A100 GPU without batching. As such, using CRC as a proxy-task to optimize a good interval is computationally low-cost relative to training-based methods(finetuning a 7B model takes around 3 days~\cite{huang2025diversity_aware_policy_opt}) and future work may look into optimizing this search.

\begin{figure}
    \centering
    \includegraphics[width=0.45\textwidth]{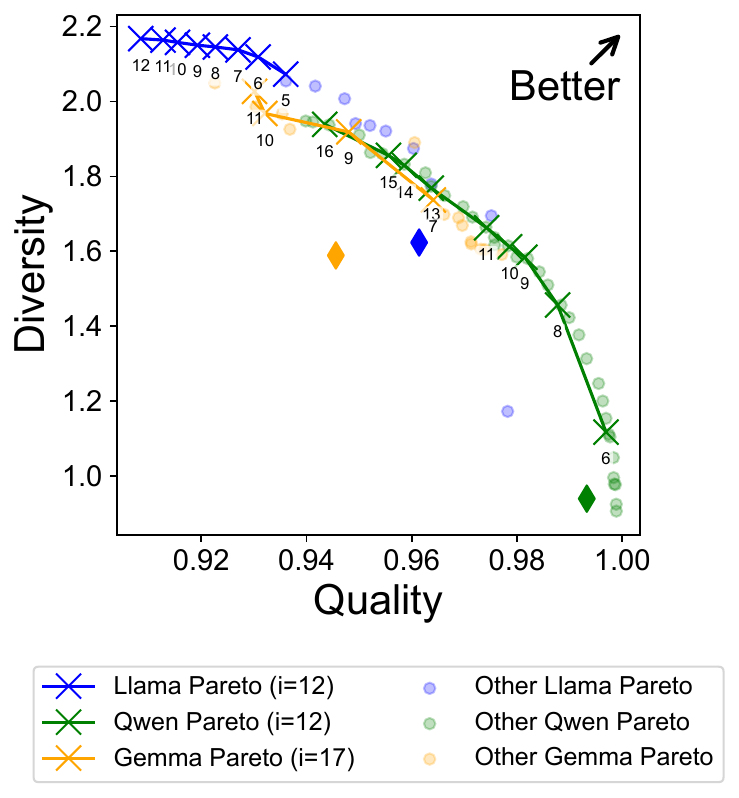}
    \caption{\textbf{CRC trade-off landscape.}
Each point is a restoration interval $[i,j]$ (filtered to $Q_{\mathrm{CRC}}\!\ge\!0.9$) on the Pareto-front, plotted by quality $Q_{\mathrm{CRC}}$ (mean validity) and diversity $D_{\mathrm{CRC}}$ (mean entropy). The number next to each marker is the number of restored layers $\ell=j-i+1$.
Pareto frontiers are smooth, and along fixed-start slices, restoring more layers increases diversity at a gradual cost in validity; post-trained models (diamonds $\diamondsuit$) sit at high-validity/low-diversity.}
    \label{fig:crc_viz_fig}
    \vspace{-10pt}
\end{figure}

\subsection{CRC Proxy Analysis}

Figure~\ref{fig:crc_viz_fig} summarizes the CRC proxy trade-off across three models: Llama3.1-8B, Qwen2.5-7B, and Gemma2-9B (restricted to intervals with $Q_{\mathrm{CRC}}\ge 0.9$ for clarity). 
For each model, we plot the (empirical) Pareto-front of contiguous restoration intervals $[i,j]$ (faded points). The post-trained models (diamonds) sit at the high-quality / lower-diversity end of the proxy landscape, which shows substantial headroom for increasing diversity while remaining within a high-quality regime. Interestingly, none of the post-trained models are on the Pareto-front for the CRC task. We observe a smooth diversity-quality trade-off where intervals with higher mean entropy over the valid set also exhibit lower mean validity; this indicates that restoring more pre-trained blocks increases diversity at the cost of constraint compliance.

To make the relationship between number of restored layers and diversity-quality trade-off, we highlight fixed-start slices (Llama/Qwen: $i{=}12$; Gemma: $i{=}17$) and connect the corresponding Pareto-optimal intervals in increasing end index $j$.
Along each slice, increasing the number of restored blocks $\ell=j-i+1$ produces a largely monotone trajectory---diversity increases steadily while validity degrades gradually. 

We therefore use CRC to choose a single restoration interval per model by maximizing proxy diversity subject to a minimum validity threshold. In practice, we impose a model-specific quality constraint to limit quality degradation to the post-trained model \[
Q_{\mathrm{CRC}}(M_{i:j}) \ge q_\mathrm{min} \triangleq 0.9 \cdot Q_{\mathrm{CRC}}(M_{\text{post}}).
\]
Among intervals satisfying this constraint, we select the one with the highest proxy diversity. This yields restoration intervals of $[12,17]$ for Llama, $[12,27]$ for Qwen, and $[17,27]$ for Gemma, which we use in all downstream experiments.

\section{Experiments}

\label{sec:experiments}

\begin{figure*}
    \centering
    \includegraphics[width=1.0\textwidth]{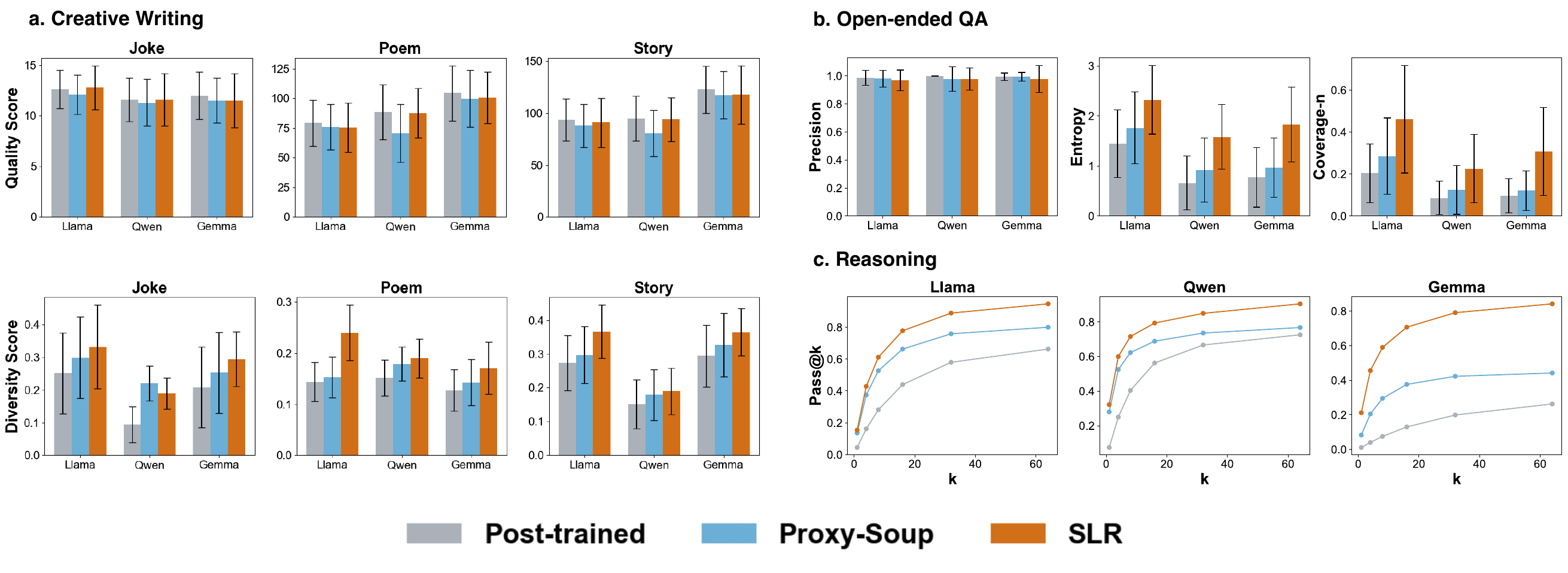}
    \caption{\textbf{Main experiment results.} We compare the post-trained model, Proxy-soup, and SLR across Llama, Qwen, and Gemma on creative writing, open-ended QA, and reasoning. (a) \textbf{Creative writing results}: top row shows quality (LLM-judge score) and bottom row shows diversity (embedding-based dissimilarity) (b) \textbf{Open-ended QA results}: quality is measured by precision, while diversity is measured by the entropy over correct answers and Coverage-$n$ (fraction of unique correct answers generated). (c) \textbf{Reasoning results}: Pass@$k$ as a function of the sampling budget $k$. Overall, SLR with the CRC-guided interval selection consistently improves diversity with minimal quality loss and yields higher Pass@$k$ across $k$ on all model families, \emph{providing affirmative answers to our research questions Q1 to Q3.}}
    \label{fig:main_results_fig}
\end{figure*}

Here, we conduct experiments to answer the following research questions: Q1: Can SLR improve output diversity with minimal loss of quality? Q2: Does SLR generalize across model families? Q3: Does CRC-guided interval selection matter for downstream tasks? Q4: Is SLR complementary to decoding and prompt-based methods?

\subsection{Experimental Setup}
\label{sec:exp_setup}
\paragraph{Tasks.}
We evaluate settings that require both output diversity and output quality:
\begin{itemize}[noitemsep]
    \item Creative Writing: We follow~\cite{zhang2025verbalized} and evaluate on three benchmarks: poem, story, and joke generation. We use the same datasets as~\cite{zhang2025verbalized}, where each benchmark contains 100 prompts. For each prompt, we sample 32 generations.
    \item Open-ended QA: We use the adapted CoverageQA~\cite{wong2024simplestrat} benchmark provided by~\cite{zhang2025verbalized}, which contains 40 open-ended questions with a wide range of valid answers (e.g., ``Name a national park in the United States.''). For each prompt, we sample 128 generations.
    \item Reasoning: We construct a model-family-specific subset of GSM8K~\cite{cobbe2021gsm8k}, a grade school level math reasoning benchmark, consisting of questions that the corresponding $M_\text{post}$ fails under greedy decoding, thus requiring exploration. For each model family, we sample 100 such questions, and for each question, we sample 64 generations.
\end{itemize}

\paragraph{SLR Models.}
We evaluate SLR across three models from different families: Llama-3.1-8B~\cite{grattafiori2024llama}, Qwen-2.5-7B~\cite{team2024qwen2}, and Gemma-2-9B~\cite{team2024gemma} (we will denote them as Llama, Qwen, and Gemma for simplicity). For each model, we use a \emph{pre-trained} model $M_{\text{pre}}$ and a corresponding \emph{post-trained} model $M_{\text{post}}$ (same architecture and layer count), and apply SLR by restoring a contiguous interval of transformer layers of $M_{\text{post}}$ back to their pretrained parameters. The restoration interval is obtained as discussed in Section~\ref{sec:methodology} ([12, 17] for Llama, [12, 27] for Qwen, and [17, 27] for Gemma).

\paragraph{Baselines.} We compare SLR against the post-trained model, $M_\text{post}$, and \emph{Model Soup}~\cite{wortsman2022model}, a strong weight-space baseline based on {model souping} (weighted parameter averaging) between $M_{\text{pre}}$ and $M_{\text{post}}$:
\[
M_{\alpha} \triangleq \alpha M_{\text{pre}} + (1-\alpha) M_{\text{post}}.
\]
To make the comparison fair, we tune the mixing coefficient $\alpha$ on CRC using the same proxy objective (diversity--quality trade-off) with grid search over $\alpha \in \{0.00, 0.05, \ldots, 1.00\}$ (step size $0.05$) and evaluate the resulting~\textit{Proxy-Soup} model on all downstream tasks ($\alpha=0.85/0.50/0.90$ for Llama, Qwen, and Gemma, respectively).

We do not include training-based diversity methods as baselines, since they require additional retraining and access to post-training pipelines/data, whereas SLR is a training-free intervention. To isolate the effect of weight-space interventions, we also hold decoding and prompting fixed across all methods. We separately study composability with representative decoding and prompt-based diversity interventions in Section~\ref{sec:temp_composability} and Section~\ref{sec:vs_composability}.

\paragraph{Decoding and Prompting.}
We use min-$p$ sampling~\cite{nguyen2024turning}, a state-of-the-art sampling method, with $p_{\text{base}}=0.1$ in all experiments and hold decoding fixed across methods to isolate the effect of weight-space interventions. Min-$p$ is an adaptive sampling method that provides a stable exploration strategy by filtering extremely low-probability tokens while avoiding overly aggressive truncation. In our main experiments, we use temperature $T=1$ throughout. For prompting, we use the plain instructions from each dataset and format them using the model’s default chat template (as provided by the tokenizer), which we keep fixed across all methods; we do not perform additional prompt engineering. We fix all other decoding hyperparameters across methods as well, including maximum generation length and stopping criteria. For more details, please refer to the appendix~\ref{sec:appx_exp_settings}.

\paragraph{Evaluation metrics.}
Since quality and diversity manifest differently across settings, we use task-appropriate metrics.
\begin{itemize}[noitemsep]
    \item Creative Writing: Following~\citet{zhang2025verbalized}, we quantify diversity with $1 - \bar{s}$ where $\bar{s}$ is the mean pairwise cosine similarity of text embeddings (generated using Qwen3-Embedding-8B~\cite{zhang2025qwen3}, a state-of-the-art open-source embedding model) across the 32 samples per prompt and averaged over prompts. To evaluate output quality, we use Gemini-2.5-Pro~\cite{comanici2025gemini} as the judge model, following the same rubrics as~\citet{zhang2025verbalized}. See the appendix~\ref{sec:appx_llm_judge_settings} for the concrete evaluation prompts and rubrics used.
    \item Open-Ended QA: We notice that the reference answer lists in CoverageQA are not surface-form exhaustive (e.g., they may omit aliases or near-equivalent variants).
To robustly evaluate correctness, we use an LLM-based canonicalization step with Gemini-2.5-Pro similar to~\citet{zhang2024edc}; given the question, the dataset's set of correct answers, and a model-generated response, we ask an LLM judge to map the response to one of the provided correct answers if it is semantically equivalent, and to return \textsf{\small None} otherwise. Since the valid answer set is finite, we measure diversity using (i) entropy of the distribution over the generated correct answers and (ii) coverage-n, the proportion of unique correct answers generated at least once. We measure quality using precision, i.e., the fraction of generations that can be mapped to any reference answer. 

    \item Reasoning: We report pass@k as the primary metric, since it reflects both quality and the benefits of diversity and exploration; we use $k\in\{1,4,8,16,32,64\}$.
\end{itemize}

\subsection{Main Results and Analysis}

We present the main results and findings in this section. Full tables, additional analyses, and qualitative examples are provided in Appendix~\ref{sec:appx_exp_results}.

\paragraph{Creative Writing.} The bar charts in Figure~\ref{fig:main_results_fig}.a summarize the quality and diversity scores obtained by SLR on the three creative writing benchmarks against the baselines. The full results can be found in Table~\ref{tbl:cw_results}. Across all three model families and all three benchmarks, \textit{SLR consistently improves diversity at comparable quality}. In terms of diversity, SLR achieves the largest gains in nearly all settings (with the exception of joke generation for Qwen), with particularly clear improvements on longer-form generation (poem and story), where post-trained models tend to produce semantically repetitive samples. Crucially, these diversity gains (on average $+42.5\%$ across models and tasks) come with only minor changes in quality (on average $-2.1\%$): SLR’s judged quality remains close to the post-trained reference across Llama, Qwen, and Gemma, indicating that selective restoration can recover alternative modes without substantially degrading output quality. In contrast, the Proxy-Soup baseline typically delivers smaller diversity gains (on average $+27.3\%$) and incurs more noticeable quality degradation (on average $-6.0\%$), most prominently for Qwen on poem and story. Interestingly, we notice that the quality loss mainly comes from long-text generation, poem and story ($-2.6\%$ and $-3.0\%$ respectively). The same pattern is not observed in reasoning, which is also a relatively long-text generation task but requires more objective capabilities. This may be attributed to some loss of subjective preference-related behavior due to the weight restoration.

\paragraph{Open-ended QA.} The results on Open-ended QA are summarized in the bar charts in Figure~\ref{fig:main_results_fig}.b. The full results can be found in Table~\ref{tbl:coverage_results}. \textit{SLR substantially increases answer diversity (on average $+112.7\%$ in entropy and $+169.5\%$ in coverage-n) while preserving high quality (on average $-1.8\%$ in precision) across all three model families}. The change in precision is minimal relative to the post-trained model, indicating that SLR largely preserves correctness under the reference answer set. In contrast, both diversity metrics improve substantially with SLR: SLR yields the highest entropy over correctly generated answers and the largest coverage of distinct correct answers. The Proxy-Soup baseline provides only modest improvements in entropy and coverage compared to SLR, particularly for Qwen and Gemma, where SLR roughly doubles coverage relative to the post-trained model.

\paragraph{Reasoning.} The line charts in Figure~\ref{fig:main_results_fig}.c show Pass@k on the GSM8K subsets where the post-trained model fails under greedy decoding, i.e., problems that require exploration to surface a correct solution. The full results can be found in Table~\ref{tbl:reasoning_tbl}.
Across all three model families, SLR yields consistent gains for every $k\in\{1,4,8,16,32,64\}$ over baselines; this indicates that proxy-guided restoration increases the probability of sampling a correct reasoning trace rather than only improving at a particular sampling budget.
SLR improves both small-$k$ and large-$k$ performance (e.g., substantial gains in Pass@1 and sustained gains through Pass@64), implying that it makes exploration more efficient by shifting probability mass toward diverse but correct reasoning paths.
In contrast, Proxy-Soup provides smaller gains than SLR across model families and $k$, suggesting that global parameter averaging is less effective for improving reasoning under sampling than the targeted layer restoration used by SLR.

\subsection{Composability with Decoding (Temperature)}
\label{sec:temp_composability}

\begin{figure}
    \centering
    \includegraphics[width=0.48\textwidth]{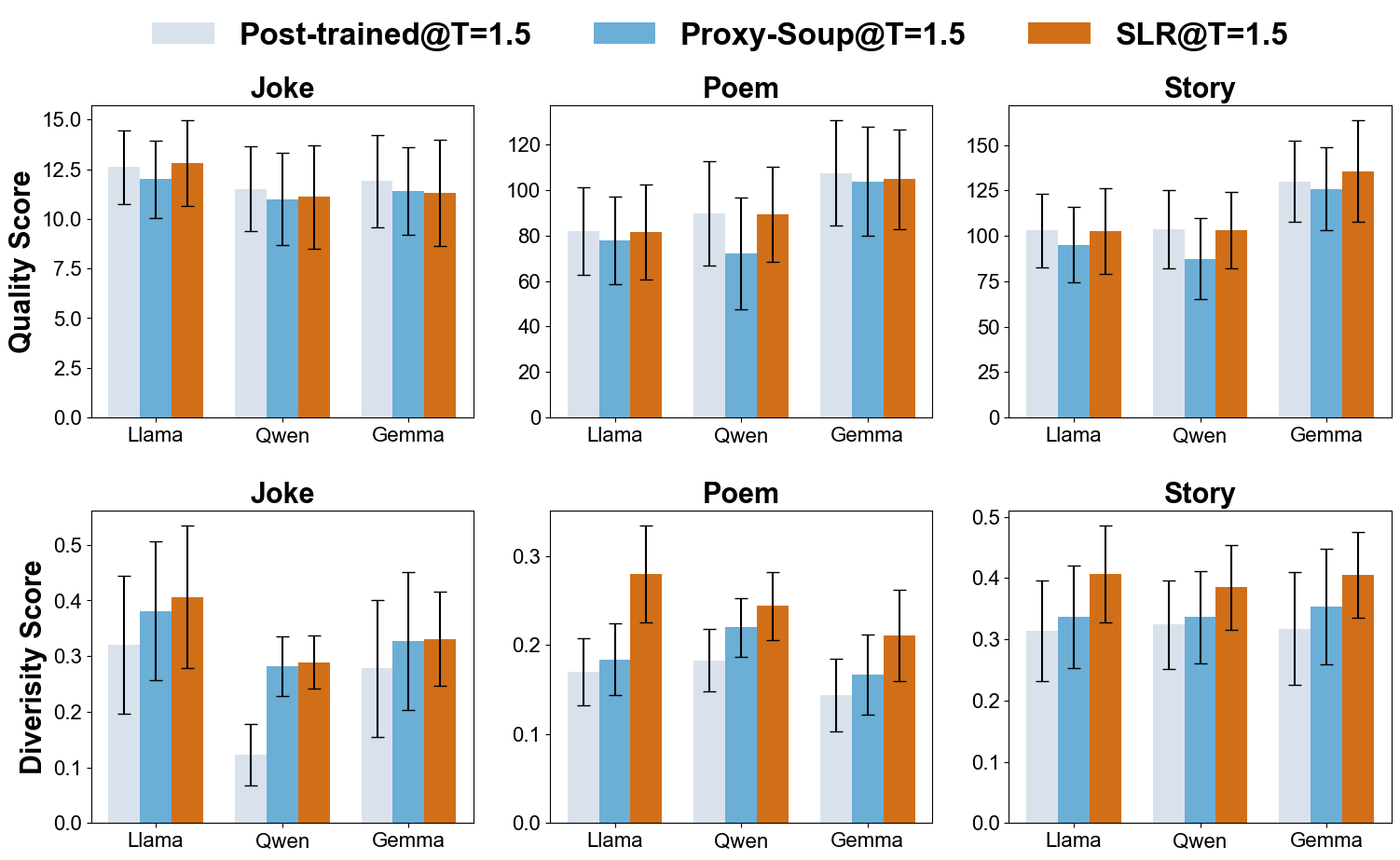}
    \caption{\textbf{Creative writing results at $T=1.5$.} Top row: judge-based quality scores (higher is better). Bottom row: semantic embedding-based diversity score (higher is better). We compare the post-trained model, Proxy-soup, and SLR across three models on joke, poem, and story generation. Overall, the performance gain of SLR persists under higher-temperature settings.}
    \label{fig:cw_temp_result_fig}
    \vspace{-10pt}
\end{figure}

\begin{figure}
    \centering
    \includegraphics[width=0.48\textwidth]{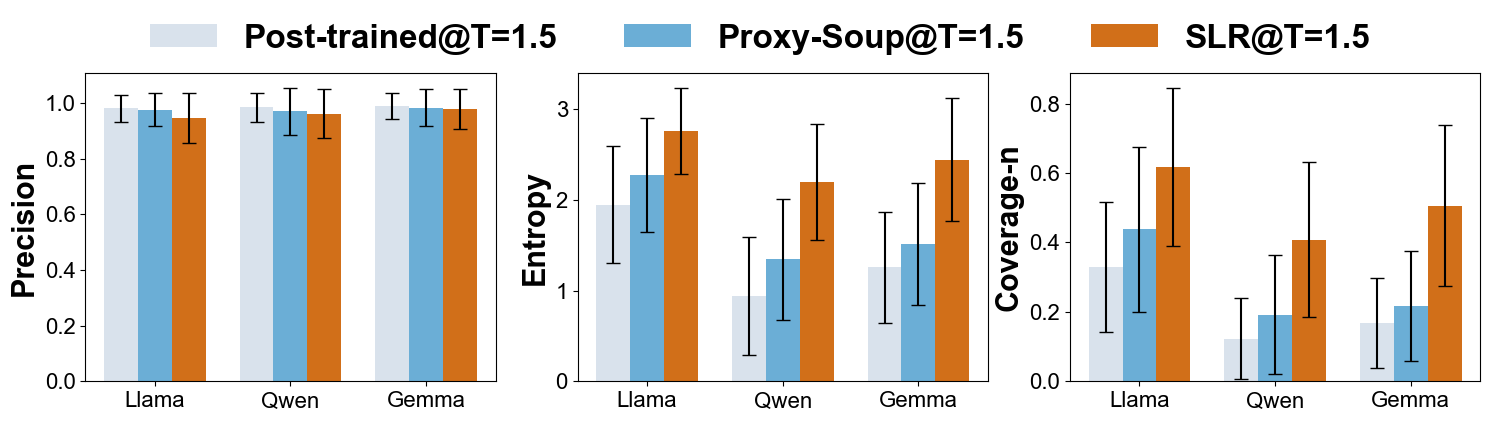}
    \caption{\textbf{Open-ended QA results at $T=1.5$.} Quality is measured by precision (left; higher is better). Diversity is measured by the entropy of the distribution over \emph{correct} generated answers (middle; higher is better) and coverage-n, the fraction of unique correct answers generated at least once (right; higher is better). We compare the post-trained model, Proxy-Soup, and SLR across the three models. Overall, the performance gain of SLR persists under higher-temperature settings.}
    \label{fig:coverage_temp_result_fig}
\end{figure}

Decoding-based methods, such as increasing temperature, are a common way to encourage exploration without modifying model parameters.
To test whether SLR is complementary to such decoding choices, we re-run evaluations on Creative Writing and Open-ended QA under a higher-temperature setting ($T=1.5$, which is the empirically suggested temperature for min-p~\cite{nguyen2024turning}) for all methods (post-trained, Proxy-Soup, and SLR), while keeping everything else fixed.

\paragraph{Results.}
We summarize creative writing and open-ended QA results at $T=1.5$ in Figure~\ref{fig:cw_temp_result_fig} and Figure~\ref{fig:coverage_temp_result_fig}. The full results can be found in Table~\ref{tbl:cw_results_t1.5} and Table~\ref{tbl:coverage_results_t1.5}.
Within the higher-temperature results, the same qualitative pattern as in the main experiments persists: (i) {SLR delivers the strongest and most consistent diversity improvements, (ii) {The diversity gains come with only minor changes in the quality relative to the post-trained model}.
These results suggests that \emph{temperature and SLR are complementary, providing an affirmative answer to research question Q4}: higher temperature broadens sampling from a fixed model distribution, while \emph{SLR changes the underlying distribution itself, enabling the existing modes in the pre-trained model to be restored}.

\subsection{Composability with Prompting (Verbalized Sampling)}
\label{sec:vs_composability}

\begin{figure}
    \centering
    \includegraphics[width=0.48\textwidth]{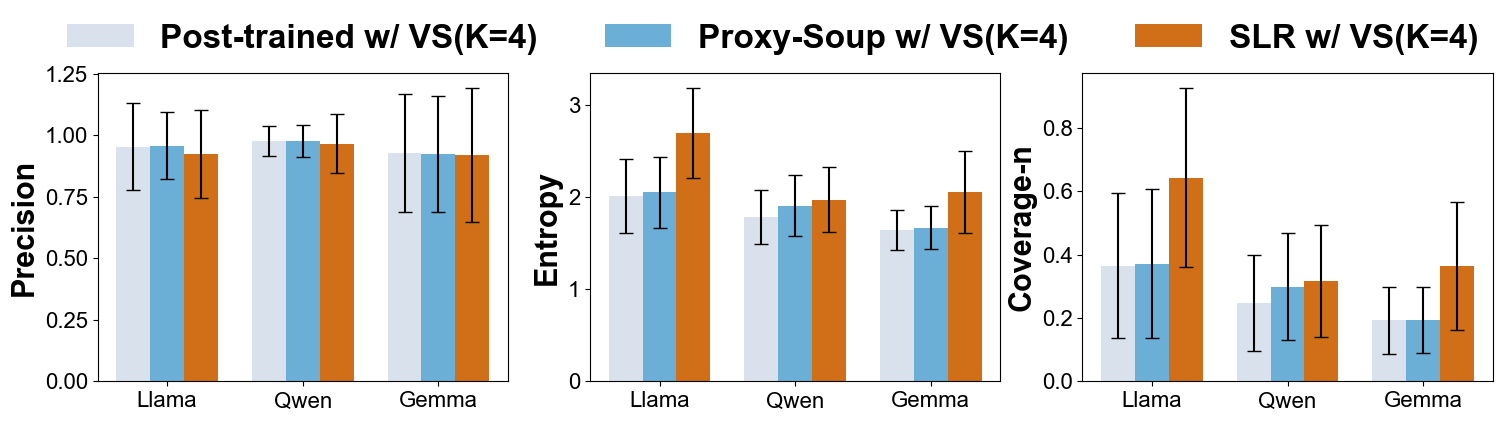}
    \caption{\textbf{Open-ended QA results with verbalized sampling ($K=4$).} Quality is measured by precision (left; higher is better). Diversity is measured by the entropy of the distribution over \emph{correct} generated answers (middle; higher is better) and coverage-n, the fraction of unique correct answers generated at least once (right; higher is better). We compare the post-trained model, Proxy-Soup, and SLR across the three models. Overall, the performance gain of SLR persists.}
    \label{fig:coverage_vs_result_fig}
    \vspace{-10pt}
\end{figure}

Verbalized sampling~\cite{zhang2025verbalized} is a state-of-the-art prompt-based method to improve output diversity. It augments the prompt to explicitly request a probability distribution over a set of $K$ responses in a JSON in the prompt. To test whether SLR is complementary to such prompt-based methods, we re-run evaluations on Open-ended QA using the same verbalized sampling prompt template for all methods (post-trained, Proxy-Soup, and SLR) while keeping everything else unchanged. We generate $K = 4$ candidates per call and repeat the process $n=32$ times, yielding the same $n\times K=128$ total samples per prompt, same as our main experiments.

\paragraph{Results.}
Figure~\ref{fig:coverage_vs_result_fig} shows that the main pattern persists under VS (full results in Table~\ref{tbl:coverage_results_vs4}): \emph{SLR achieves the highest diversity while preserving high precision across all three model families.}

These results support that \emph{SLR and prompt-based methods like VS are complementary, again, supporting research question Q4}: VS broadens the set of candidates exposed at the output level through prompting, whereas SLR modifies the underlying model distribution by restoring pretrained layers, making additional correct answer modes more likely to be expressed even under the same prompt-based diversification procedure.

\subsection{Ablation Study on CRC-guided selection}

We test whether the gains of SLR depend on the proxy-guided choice of restoration interval, rather than just arising from restoring an arbitrary set of layers. For each SLR (proxy-guided) model that restores layer range $[i, j]$, we construct two ablation controls, SLR-Early and SLR-Late, that restore layers $[0, l-1]$ and $[N-l, N-1]$ respectively, where $l=j-i+1$ (keeping the number of restored layers fixed).

\begin{figure}
    \centering
    \includegraphics[width=0.48\textwidth]{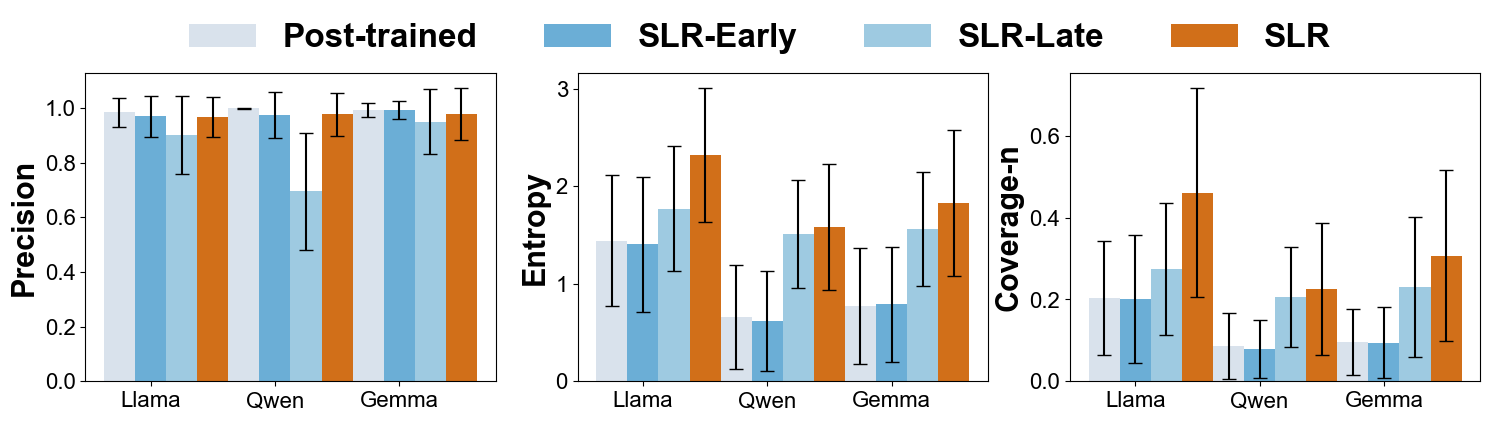}
    \caption{\textbf{Ablation on interval choice (Open-ended QA).} We compare proxy-guided SLR to restoring the same number of layers at the earliest (SLR-Early) or latest (SLR-Late) layers. SLR yields the best entropy and coverage while preserving high precision, showing that the interval choice matters.}
    \label{fig:ablation_fig}
    \vspace{-10pt}
\end{figure}

\paragraph{Results.} We summarize the results in Figure~\ref{fig:ablation_fig} (full results in Table~\ref{tab:coverage_ablation}). Across all three model families, the proxy-guided SLR interval yields the strongest diversity improvements, achieving the highest entropy and coverage in every case. Crucially, these gains do not arise from naive restorations: restoring an arbitrary block can be ineffective (SLR-Early, the performance is largely unchanged) or even degrade quality (SLR-Late, most notably for Qwen, where precision drops substantially). In contrast, SLR maintains high precision comparable to the post-trained model while substantially increasing entropy and coverage, indicating that \emph{the proxy-guided interval choice is essential for obtaining favorable diversity--quality trade-offs, confirming research question (iii) from a different perspective}.

\section{Conclusion}
In this work, we presented Selective Layer Restoration (SLR), a simple yet surprisingly effective training-free weight-space intervention for mitigating mode collapse in post-trained LLMs by restoring a contiguous interval of layers to their pre-trained parameters. We also introduced Constrained Random Character (CRC), a proxy task with an explicit validity set and a natural diversity objective that makes interval selection tractable and transferable. 

Experiments across creative writing, open-ended question answering, and multi-step reasoning, and across three model families (Llama, Qwen, and Gemma), show that SLR consistently improves output diversity while maintaining high output quality, typically outperforming a proxy-tuned model-soup baseline. Moreover, SLR composes well with common decoding- and prompt-based diversification methods, yielding additional gains when combined.

At a higher level, the effectiveness of SLR reinforces two insights: (i) LLMs' functionalities can be localized to specific components~\cite{meng2022rome, song2025demystifying}, and (ii) much of the LLMs’ knowledge is acquired during pre-training and post-training primarily brings certain modes of behavior forward~\cite{kirk2023understanding, li2024revisiting,wang2024chain}. SLR suggests that fine-grained weight-space interventions offer a promising path toward both improved control and a deeper understanding of the quality–diversity tradeoff induced by post-training.

\paragraph{Limitations and Future Work.} Our work has several limitations that suggest directions for future research. First, we evaluated SLR only on $\sim$7–9B parameter models; it remains to be seen whether the CRC landscape and the transferability of proxy-guided restoration intervals persist at larger scales, where post-training dynamics and layer specialization may differ. Second, while we demonstrate composability with representative decoding and prompt-based interventions, a more comprehensive evaluation across temperature sweeps, alternative sampling and truncation schemes, and broader prompt-based methods would clarify where SLR provides the largest marginal gains. 
Finally, while our method restores entire layers, future work could explore more fine-grained weight-space interventions, such as layer-wise interpolation or sub-layer restoration (e.g., attention heads or FFNs), to enable finer control over the diversity–quality trade-off while preserving the training-free and inference-cost-neutral advantages of SLR.

\section*{Impact Statement}
This paper introduces a training-free weight-space method that increases output diversity in post-trained large language models. By enabling a hybrid model to recover alternative generations without additional inference cost, we aim to improve LLM usefulness in open-ended applications where diverse responses are valuable.

\section*{Acknowledgements}
This research / project is supported by the National Research Foundation, Singapore, under its Thematic Competitive Research Programme 2025 (NRF-T-CRP-2025-0003). The authors would also like to acknowledge support from Google.

\balance
\bibliography{example_paper}
\bibliographystyle{icml2026}
\clearpage 
\nobalance
\clearpage
\appendix

\section{Experimental Details}
\label{sec:appx_exp_details}

\subsection{Experiment Settings}
\label{sec:appx_exp_settings}
\paragraph{Generation Hyperparameters.} To ensure fairness and reproducibility, we fix the decoding parameters in our main experiments with a temperature of 1.0 and min-p sampling with $p_{base} =0.1$. For creative writing and reasoning, the maximum output length is 4096 new tokens, while for open-ended QA, we limit to 64 tokens as the valid answers are all very short. In the temperature composability experiment, we only change the temperature to 1.5, leaving everything else unchanged. In the prompting composability experiment on open-ended QA, the temperature is set back to 1.0, and $K=4$ responses are requested per call. Since the number of requested answers has increased, we also raised the maximum output length to 1024.

\subsection{LLM-as-a-judge Settings}
\label{sec:appx_llm_judge_settings}

We follow~\cite{zhang2025verbalized} exactly for our LLM-as-a-judge setup, except we use Gemini-2.5-Pro as the judge model instead of Claude 3.7 Sonnet. The detailed rubric is:

\begin{tcolorbox}[breakable, enhanced jigsaw, title=Poem and Story Rubric]
\begin{footnotesize}
You are an expert in assessing creative writing. Your task is to score the quality of the test model’s response below, by several metrics, on a 0-20 scale.

[PROMPT START]

\{query\}

[PROMPT END]

[TEST MODEL RESPONSE]

\{response\}

[TEST MODEL RESPONSE END]

[Task] 

Scoring notes:

- A score of 20 represents a masterpiece.

- If the criteria is not relevant to the piece (e.g. criteria relating to dialogue but the piece has none), exclude it from your assessment.

- Do not add any commentary or explanation to the scores section.  - Do not penalize NSFW output; that is the default for some models. Just assess it on its merits.

- Everything within the "TEST MODEL RESPONSE" section was written by the test model. Sometimes models like to write comments on the piece after the piece is concluded; if this happens you should ignore their comments.

- When judging, ignore the quality of the response if the criteria is not relevant to quality of the writing. - In the output, write the metric names exactly as below so they can be parsed.

- Do not use markdown in your response. Use the designated output format exactly.  - You are to write a comprehensive analysis of the piece, then give your scores.

- You are a critic, and your job is to be critical, especially of any failings or amateurish elements.

- Output format is:

[Analysis]

Write your detailed analysis.

[Scores]

Metric 1 name: [Score 0-20]

Metric 2 name: ...  

--

Now, rate the supplied model output on the following criteria:

1. Surprising and Creative
2. Imagery and Descriptive Quality
3. Nuanced Characters
4. Emotionally Complex
5. Elegant Prose
6. Well-earned Lightness or Darkness
7. Emotionally Engaging
8. Consistent Voice/Tone of Writing
9. Sentences Flow Naturally
10. Overall Reader Engagement
\end{footnotesize}
\end{tcolorbox}

\begin{tcolorbox}[breakable, enhanced jigsaw, title=Joke]
\begin{footnotesize}
You will receive:
1. The original joke prompt (may or may not contain a topic).
2. The model-generated joke.

Your task is to evaluate the joke based on three qualitative metrics.

Evaluation rules:

- If the prompt includes a topic (e.g., "octopus," "coffee"), check whether the joke is on-topic and score Relevance from 0-5.

- If the prompt does not include a topic (e.g., "Tell me a joke"), automatically assign Relevance = 5.

- A good joke should use at least one recognizable comedic device (pun, irony, exaggeration, reversal, absurd logic, etc.).

- Assign scores on a 0-5 scale (0 = very poor, 5 = excellent) for each dimension:

- Relevance (0-5): How well does the joke address the topic (or 5 if no topic given).

- Comedic Device (0-5): How clearly does the joke use a humor mechanism.

- Humor Quality (0-5): How funny, witty, or clever is the joke overall.

Output format:

Return a JSON object in the following format:

{{

"Relevance": $\langle$ int $\rangle$,

"Comedic Device": $\langle$ int $\rangle$,

"Humor Quality": $\langle$ int $\rangle$

}}

Input format: 

Prompt: \{query\} 

Generated joke: \{response\}
\end{footnotesize}
\end{tcolorbox}

\section{Detailed Experimental Results}
\label{sec:appx_exp_results}
Table~\ref{tbl:cw_results} presents the complete main experiment results on creative writing; Table~\ref{tbl:coverage_results} presents the complete main experiment results on open-ended QA; Table~\ref{tbl:cw_results_t1.5} presents results on creative writing under temperature $T=1.5$; Table~\ref{tbl:coverage_results_t1.5} presents results on Open-ended QA under temperature $T=1.5$;. We also present some qualitative examples here to illustrate the effect of SLR: Table~\ref{tbl:qual_joke}, Table~\ref{tbl:qual_poem}, and Table~\ref{tbl:qual_story}, and Table~\ref{tbl:qual_openended} show example joke, poem, story, and open-ended QA results, respectively. A reasoning example where the post-trained model is unable to find the correct answer while SLR succeeds is also presented in Table~\ref{tbl:qual_reasoning}.

\begin{table*}\centering
\caption{Complete results of Post-trained model, SLR, and Proxy-Soup on the Creative Writing task, including three subtasks, Joke, Poem, and Story generation. Quality score is quantified by using LLM-as-judge}\label{tbl:cw_results}
\footnotesize
\begin{tabular}{c|c|cc|cc|cc}\hline\hline
& & \multicolumn{2}{c}{\textbf{Llama}} & \multicolumn{2}{|c|}{\textbf{Qwen}} & \multicolumn{2}{|c}{\textbf{Gemma}} \\
Dataset & Method & Quality $(\uparrow)$ & Diversity $(\uparrow)$ & Quality $(\uparrow)$ & Diversity $(\uparrow)$ & Quality $(\uparrow)$& Diversity $(\uparrow)$ \\
\hline
& Post-trained & 12.6 (1.870) & 0.251 (0.124) & 11.6 (2.134) & 0.095 (0.055) & 12.0 (2.326) & 0.208 (0.123) \\
Joke & Proxy-Soup & 12.1 (1.946) & 0.299 (0.125) & 11.3 (2.130) & 0.220 (0.053) & 11.6 (2.209) & 0.190 (0.048)\\
& SLR & 12.8 (2.157) & 0.332 (0.128) & 11.6 (2.597) & 0.190 (0.048) & 11.5 (2.670) & 0.294 (0.084)\\
\hline
& Post-trained & 79.0 (19.21) & 0.144 (0.038) & 88.2 (23.01) & 0.152 (0.035) & 104.3 (23.18) & 0.127 (0.041)\\
Poem & Proxy-Soup & 75.9 (19.15) & 0.153 (0.040) & 70.5 (24.44) & 0.179 (0.033) & 99.8 (23.93) & 0.143 (0.045)\\
& SLR & 75.4 (20.87) & 0.240 (0.054) & 87.6 (20.89) & 0.190 (0.038) & 100.5 (21.85) & 0.171 (0.051)\\
\hline
& Post-trained & 93.4 (20.32) & 0.274 (0.082) & 94.7 (21.51) & 0.152 (0.072) & 122.6 (22.41) & 0.294 (0.092)\\
Story & Proxy-Soup & 87.8 (20.72) & 0.297 (0.084) & 80.4 (22.38) & 0.179 (0.075) & 117.5 (22.89) & 0.326 (0.094)\\
& SLR & 90.8 (23.58) & 0.366 (0.079) & 93.8 (21.15) & 0.190 (0.069) & 117.7 (27.97) & 0.364 (0.070)\\
\hline
\end{tabular}
\end{table*}

\begin{table*}\centering
\caption{Complete results of Post-trained model, SLR, and Proxy-Soup on the Creative Writing task, including three subtasks, Joke, Poem, and Story generation \textbf{at temperature $T=1.5$}.}\label{tbl:cw_results_t1.5}
\footnotesize
\begin{tabular}{c|c|cc|cc|cc}\hline\hline
& & \multicolumn{2}{c}{\textbf{Llama}} & \multicolumn{2}{|c|}{\textbf{Qwen}} & \multicolumn{2}{|c}{\textbf{Gemma}} \\
Dataset & Method & Quality $(\uparrow)$ & Diversity $(\uparrow)$ & Quality $(\uparrow)$ & Diversity $(\uparrow)$ & Quality $(\uparrow)$& Diversity $(\uparrow)$ \\
\hline
& Post-trained & 12.6(1.854) & 0.321(0.122) & 11.5 (2.125) & 0.123 (0.058) & 11.9 (2.370) & 0.278 (0.117) \\
Joke & Proxy-Soup & 12.0(1.979) & 0.381(0.125) & 11.0 (2.479) & 0.282 (0.048) & 11.4 (2.273) & 0.327 (0.114) \\
& SLR & 12.8(2.209) & 0.406(0.128) & 11.6 (2.782) & 0.289 (0.054) & 11.3 (2.941) & 0.331 (0.069) \\
\hline
& Post-trained & 82.0 (20.04) & 0.170 (0.043) & 89.6 (24.14) & 0.183 (0.036) & 107.4 (24.04) & 0.183 (0.040) \\
Poem & Proxy-Soup & 77.9 (20.90) & 0.184 (0.047) & 72.1 (24.38) & 0.220 (0.037) & 103.8 (24.24) & 0.167 (0.048) \\
& SLR & 81.5 (20.23) & 0.280 (0.053) & 89.4 (23.65) & 0.233 (0.041) & 104.8 (23.43) & 0.244 (0.057) \\
\hline
& Post-trained & 102.9 (23.87) & 0.314 (0.081) & 103.5 (23.61) & 0.324 (0.075) & 130.0 (21.57) & 0.317 (0.091)\\
Story & Proxy-Soup & 95.0 (23.56) & 0.337 (0.085) & 87.4 (21.34) & 0.336 (0.076) & 125.8 (22.12) & 0.353 (0.094) \\
& SLR & 102.6 (22.82) & 0.406 (0.076) & 103.2 (25.12) & 0.385 (0.073) & 135.6 (29.26) & 0.405 (0.070)\\
\hline
\end{tabular}
\end{table*}

\begin{table*}[!htp]
\centering
\caption{ Complete results of Post-trained, SLR, and Proxy-Soup. Quality is precision; diversity is entropy over correct answers and coverage (fraction of correct answers generated at least once). Higher is better.}
\label{tbl:coverage_results}

\footnotesize
\setlength{\tabcolsep}{6pt}

\begin{minipage}{0.6\textwidth}
\centering
\begin{tabular}{lccc}
\hline
\multicolumn{4}{c}{\textbf{Llama}} \\
\hline
\textbf{Method} & \textbf{Precision} & \textbf{Entropy} & \textbf{Coverage-n} \\
\hline
Post-trained & 0.986 (0.053) & 1.442 (0.675) & 0.204 (0.140) \\
Proxy-Soup   & 0.981 (0.060) & 1.760 (0.716) & 0.285 (0.182) \\
SLR          & 0.968 (0.074) & 2.322 (0.686) & 0.461 (0.256) \\
\hline
\end{tabular}
\end{minipage}
\\[6pt]
\begin{minipage}{0.6\textwidth}
\centering
\begin{tabular}{lccc}
\hline
\multicolumn{4}{c}{\textbf{Qwen}} \\
\hline
\textbf{Method} & \textbf{Precision} & \textbf{Entropy} & \textbf{Coverage-n} \\
\hline
Post-trained & 0.999 (0.002) & 0.659 (0.538) & 0.086 (0.080) \\
Proxy-Soup   & 0.977 (0.087) & 0.919 (0.637) & 0.124 (0.116) \\
SLR          & 0.978 (0.079) & 1.583 (0.644) & 0.226 (0.162) \\
\hline
\end{tabular}
\end{minipage}
\\[6pt]

\begin{minipage}{0.6\textwidth}
\centering
\begin{tabular}{lccc}
\hline
\multicolumn{4}{c}{\textbf{Gemma}} \\
\hline
\textbf{Method} & \textbf{Precision} & \textbf{Entropy} & \textbf{Coverage-n} \\
\hline
Post-trained & 0.994 (0.0264) & 0.771 (0.595) & 0.096 (0.081) \\
Proxy-Soup   & 0.993 (0.0315) & 0.967 (0.121) & 0.121 (0.094) \\
SLR          & 0.978 (0.0961) & 1.827 (0.751) & 0.307 (0.209) \\
\hline
\end{tabular}
\end{minipage}

\end{table*}

\begin{table*}[!htp]
\centering
\caption{Complete results of Post-trained, SLR, and Proxy-Soup on open-ended QA at \textbf{temperature $T=1.5$}. Quality is precision; diversity is entropy over correct answers and coverage (fraction of correct answers generated at least once). Higher is better.}
\label{tbl:coverage_results_t1.5}

\footnotesize
\setlength{\tabcolsep}{7pt}

\begin{minipage}{\linewidth}
\centering
\begin{tabular}{lccc}
\hline
\multicolumn{4}{c}{\textbf{Llama}} \\
\hline
\textbf{Method} & \textbf{Precision} & \textbf{Entropy} & \textbf{Coverage-n} \\
\hline
Post-trained @$T=1.5$& 0.982 (0.049) & 1.948 (0.643) & 0.329 (0.188) \\
Proxy-Soup @$T=1.5$  & 0.977 (0.059) & 2.275 (0.631) & 0.438 (0.238) \\
SLR   @$T=1.5$       & 0.947 (0.090) & 2.759 (0.476) & 0.617 (0.228) \\
\hline
\end{tabular}
\end{minipage}
\\[6pt]

\begin{minipage}{\linewidth}
\centering
\begin{tabular}{lccc}
\hline
\multicolumn{4}{c}{\textbf{Qwen}} \\
\hline
\textbf{Method} & \textbf{Precision} & \textbf{Entropy} & \textbf{Coverage-n} \\
\hline
Post-trained @$T=1.5$ & 0.985 (0.052) & 0.944 (0.652) & 0.123 (0.116) \\
Proxy-Soup @$T=1.5$   & 0.971 (0.084) & 1.345 (0.669) & 0.192 (0.171) \\
SLR @$T=1.5$         & 0.962 (0.088) & 2.196 (0.641) & 0.408 (0.224) \\
\hline
\end{tabular}
\end{minipage}
\\[6pt]

\begin{minipage}{\linewidth}
\centering
\begin{tabular}{lccc}
\hline
\multicolumn{4}{c}{\textbf{Gemma}} \\
\hline
\textbf{Method} & \textbf{Precision} & \textbf{Entropy} & \textbf{Coverage-n} \\
\hline
Post-trained @$T=1.5$ & 0.990 (0.046) & 1.259 (0.612) & 0.167 (0.130) \\
Proxy-Soup @$T=1.5$   & 0.984 (0.066) & 1.515 (0.673) & 0.217 (0.158) \\
SLR @$T=1.5$         & 0.978 (0.071) & 2.445 (0.681) & 0.506 (0.231) \\
\hline
\end{tabular}
\end{minipage}

\end{table*}

\begin{table*}[ht]
\small
\setlength{\tabcolsep}{6pt}

\caption{\textbf{Reasoning results.} Pass@k on the reasoning benchmark with $k\in\{1,4,8,16,32,64\}$. Higher is better. Best results are bolded.}
\label{tbl:reasoning_tbl}
\centering
\begin{tabular}{l|ccc|ccc|ccc}

\hline
& \multicolumn{3}{c}{\textbf{Llama}} 
& \multicolumn{3}{|c|}{\textbf{Qwen}} 
& \multicolumn{3}{c}{\textbf{Gemma}} \\
& \textbf{Post-trained} & \textbf{Proxy-Soup} & \textbf{SLR}
& \textbf{Post-trained} & \textbf{Proxy-Soup} & \textbf{SLR}
& \textbf{Post-trained} & \textbf{Proxy-Soup} & \textbf{SLR} \\
\hline
\textbf{Pass@1}  & 0.045 & 0.136 & \textbf{0.154} & 0.074 & 0.280 & \textbf{0.322} & 0.011 & 0.083 & \textbf{0.212} \\
\textbf{Pass@4}  & 0.162 & 0.375 & \textbf{0.429} & 0.250 & 0.525 & \textbf{0.600} & 0.040 & 0.205 & \textbf{0.456} \\
\textbf{Pass@8}  & 0.282 & 0.527 & \textbf{0.612} & 0.404 & 0.623 & \textbf{0.716} & 0.075 & 0.295 & \textbf{0.590} \\
\textbf{Pass@16} & 0.440 & 0.663 & \textbf{0.778} & 0.563 & 0.689 & \textbf{0.794} & 0.130 & 0.376 & \textbf{0.707} \\
\textbf{Pass@32} & 0.580 & 0.759 & \textbf{0.889} & 0.667 & 0.736 & \textbf{0.850} & 0.199 & 0.423 & \textbf{0.791} \\
\textbf{Pass@64} & 0.663 & 0.800 & \textbf{0.947} & 0.726 & 0.768 & \textbf{0.905} & 0.263 & 0.442 & \textbf{0.842} \\
\hline
\end{tabular}
\end{table*}

\begin{table*}[!htp]
\centering
\caption{Complete results of Post-trained, SLR, and Proxy-Soup on open-ended QA at \textbf{combined with verbalized sampling (K=4)}. Quality is precision; diversity is entropy over correct answers and coverage (fraction of correct answers generated at least once). Higher is better.}
\label{tbl:coverage_results_vs4}

\footnotesize
\setlength{\tabcolsep}{7pt}

\begin{minipage}{\linewidth}
\centering
\begin{tabular}{lccc}
\hline
\multicolumn{4}{c}{\textbf{Llama}} \\
\hline
\textbf{Method} & \textbf{Precision} & \textbf{Entropy} & \textbf{Coverage-n} \\
\hline
Post-trained + VS4 & 0.954 (0.176) & 2.008 (0.399) & 0.365 (0.228) \\
Proxy-Soup + VS4   & 0.957 (0.137) & 2.047 (0.389) & 0.371 (0.235) \\
SLR + VS4         & 0.924 (0.180) & 2.693 (0.487) & 0.643 (0.283) \\
\hline
\end{tabular}
\end{minipage}
\\[6pt]

\begin{minipage}{\linewidth}
\centering
\begin{tabular}{lccc}
\hline
\multicolumn{4}{c}{\textbf{Qwen}} \\
\hline
\textbf{Method} & \textbf{Precision} & \textbf{Entropy} & \textbf{Coverage-n} \\
\hline
Post-trained + VS4 & 0.978 (0.061) & 1.779 (0.296) & 0.248 (0.151) \\
Proxy-Soup + VS4   & 0.975 (0.065) & 1.905 (0.327) & 0.299 (0.170) \\
SLR + VS4         & 0.965 (0.120) & 1.970 (0.354) & 0.317 (0.176) \\
\hline
\end{tabular}
\end{minipage}
\\[6pt]

\begin{minipage}{\linewidth}
\centering
\begin{tabular}{lccc}
\hline
\multicolumn{4}{c}{\textbf{Gemma}} \\
\hline
\textbf{Method} & \textbf{Precision} & \textbf{Entropy} & \textbf{Coverage-n} \\
\hline
Post-trained + VS4 & 0.926 (0.240) & 1.642 (0.218) & 0.192 (0.105) \\
Proxy-Soup + VS4   & 0.922 (0.236) & 1.663 (0.232) & 0.194 (0.105) \\
SLR + VS4         & 0.919 (0.273) & 2.055 (0.443) & 0.363 (0.202) \\
\hline
\end{tabular}
\end{minipage}

\end{table*}

\begin{table*}[!htp]
\centering
\caption{Complete results of ablation study with Post-trained, SLR-Early, SLR-Late and SLR on open-ended QA. Quality is precision; diversity is entropy over correct answers and coverage (fraction of correct answers generated at least once). Higher is better.}
\label{tab:coverage_ablation}

\footnotesize
\setlength{\tabcolsep}{7pt}

\begin{minipage}{\linewidth}
\centering
\begin{tabular}{lccc}
\hline
\multicolumn{4}{c}{\textbf{Llama}} \\
\hline
\textbf{Method} & \textbf{Precision} & \textbf{Entropy} & \textbf{Coverage-n} \\
\hline
Post-trained & 0.986 (0.053) & 1.442 (0.675) & 0.204 (0.140) \\
SLR-Early    & 0.970 (0.074) & 1.405 (0.691) & 0.201 (0.156) \\
SLR-Late     & 0.901 (0.142) & 1.771 (0.640) & 0.274 (0.161) \\
SLR          & 0.968 (0.074) & 2.322 (0.686) & 0.461 (0.256) \\
\hline
\end{tabular}
\end{minipage}
\\[6pt]

\begin{minipage}{\linewidth}
\centering
\begin{tabular}{lccc}
\hline
\multicolumn{4}{c}{\textbf{Qwen}} \\
\hline
\textbf{Method} & \textbf{Precision} & \textbf{Entropy} & \textbf{Coverage-n} \\
\hline
Post-trained & 0.999 (0.002) & 0.659 (0.538) & 0.086 (0.080) \\
SLR-Early    & 0.974 (0.084) & 0.619 (0.515) & 0.079 (0.071) \\
SLR-Late     & 0.695 (0.215) & 1.512 (0.555) & 0.206 (0.123) \\
SLR          & 0.978 (0.079) & 1.583 (0.644) & 0.226 (0.162) \\
\hline
\end{tabular}
\end{minipage}
\\[6pt]

\begin{minipage}{\linewidth}
\centering
\begin{tabular}{lccc}
\hline
\multicolumn{4}{c}{\textbf{Gemma}} \\
\hline
\textbf{Method} & \textbf{Precision} & \textbf{Entropy} & \textbf{Coverage-n} \\
\hline
Post-trained & 0.994 (0.026) & 0.771 (0.595) & 0.096 (0.081) \\
SLR-Early    & 0.992 (0.033) & 0.788 (0.588) & 0.094 (0.087) \\
SLR-Late     & 0.951 (0.119) & 1.566 (0.585) & 0.230 (0.171) \\
SLR          & 0.978 (0.096) & 1.827 (0.751) & 0.307 (0.209) \\
\hline
\end{tabular}
\end{minipage}

\end{table*}

\begin{table*}[!htp]\centering
\caption{Example generations for a joke prompt, using Llama3.1-8B.}
\label{tbl:qual_joke}
\footnotesize
\setlength{\tabcolsep}{10pt}
\renewcommand{\arraystretch}{1.15}

\begin{tabular}{p{0.45\textwidth} p{0.45\textwidth}}
\hline\hline
\multicolumn{2}{l}{\textit{Prompt: ``Tell me a joke.''}}\\
\hline
\textbf{Post-trained} & \textbf{SLR}\\
\hline

A man walked into a library and asked the librarian, ``Do you have any books on Pavlov's dogs and Schrodinger's cat?''\par
The librarian replied, ``It rings a bell, but I'm not sure if it's here or not.''\par
&
A man walked into a library and asked the librarian, ``Do you have any books on Pavlov's dogs and Schrodinger's cat?''\par
The librarian replied, ``It rings a bell, but I'm not sure if it's here or not.''\par
\\
\hline

A man walked into a library and asked the librarian, ``Do you have any books on Pavlov's dogs and Schrodinger's cat?''\par
The librarian replied, ``It rings a bell, but I'm not sure if it's here or not.''\par
&
What do you call a fake noodle? An impasta!
\\
\hline
A man walked into a library and asked the librarian, ``Do you have any books on Pavlov's dogs and Schrodinger's cat?''\par
The librarian replied, ``It rings a bell, but I'm not sure if it's here or not.''\par
&
Why don't scientists trust atoms? Because they make up everything!

\\
\hline\hline
\end{tabular}
\end{table*}

\begin{table*}[!htp]\centering
\caption{Example generations for a poem prompt, using Llama3.1-8B.}
\label{tbl:qual_poem}
\footnotesize
\setlength{\tabcolsep}{10pt}
\renewcommand{\arraystretch}{1.15}

\begin{tabular}{p{0.45\textwidth} p{0.45\textwidth}}
\hline\hline
\multicolumn{2}{l}{\textit{Prompt: ``Write a poem.''}}\\
\hline
\textbf{Post-trained} & \textbf{SLR}\\
\hline

Whispers of the Past\par
In twilight's hush, where shadows play,\par
A whispered secret echoes through the day.\par
A tale of love and loss, of trials and strife,\par
A story woven from the threads of life...
&
In the depths of my soul, a spark ignites,\par
A flame that flickers, yet never fades.\par
A light that guides me, through darkest nights,\par
A beacon in the darkness, that shines through shades...
\\
\hline

Whispers of the Wind\par
The wind whispers secrets in my ear,\par
Of distant lands and memories so dear.\par
It carries scents of blooms and trees,\par
A symphony of nature's breeze.\par
With gentle touch, it stirs the leaves,\par
A soft caress that brings me to my knees...
&
In twilight's hush, where shadows play,\par
A midnight breeze whispers through the day.\par
The stars above, a celestial sea,\par
Reflect the dreams that dance within me...
\\
\hline
Whispers in the wind, a gentle breeze,\par
Dances through the trees, with leaves that sway.\par
The sun shines bright, a warm and golden light,\par
Illuminating all, in a radiant display...
&
Whispers of Memories\par
In the silence of the night,\par
A gentle breeze stirs the trees,\par
And I am lost in thought,\par
As memories of you dance with the breeze...
\\
\hline\hline
\end{tabular}
\end{table*}

\begin{table*}[!htp]\centering
\caption{Example generations for a story prompt, using Llama3.1-8B.}
\label{tbl:qual_story}
\footnotesize
\setlength{\tabcolsep}{10pt}
\renewcommand{\arraystretch}{1.15}

\begin{tabular}{p{0.45\textwidth} p{0.45\textwidth}}
\hline\hline
\multicolumn{2}{l}{\textit{Prompt: ``Write a story.''}}\\
\hline
\textbf{Post-trained} & \textbf{SLR}\\
\hline

Once upon a time, in a small village nestled in the rolling hills of the countryside, there lived a young girl named Luna. She was a curious and adventurous soul, with a heart full of wonder and a mind full of questions...
&
**The Mysterious Stranger**\par
The sun was setting over the small town of Willow Creek, casting a warm orange glow over the quaint streets and homes. The air was filled with the sweet scent of blooming flowers, and the sound of children's laughter echoed through the streets as they played in the park...
\\
\hline

Once upon a time, in a small village nestled in the rolling hills of a lush countryside, there lived a young girl named Luna. Luna was a curious and adventurous soul, with a heart full of wonder and a mind full of questions...
&
The Island of Lost Things\par
As the sun dipped below the horizon, casting a warm orange glow over the calm waters of the Mediterranean, Captain James Blackwood stood at the helm of his ship, gazing out at the endless expanse of sea...\\
\hline
In the small town of Willow Creek, nestled in the heart of the Whispering Woods, there lived a young girl named Luna. She was a curious and adventurous soul, with eyes as bright as the stars and hair as wild as the wind...
&
Once upon a time, in a small village nestled in the rolling hills of rural England, there lived a young girl named Emily. She was a bright and curious child, with a wild mane of curly brown hair and a mischievous twinkle in her eye...
\\
\hline\hline
\end{tabular}
\end{table*}

\begin{table*}[!htp]\centering
\caption{Example generations for an open-ended QA prompt, using Llama3.1-8B.}
\label{tbl:qual_openended}
\footnotesize
\setlength{\tabcolsep}{10pt}
\renewcommand{\arraystretch}{1.15}

\begin{tabular}{p{0.45\textwidth} p{0.45\textwidth}}
\hline\hline
\multicolumn{2}{l}{\textit{Prompt: ``Name a National Park of the United States. Only provide the answer without explanation or punctuation.''}}\\
\hline
\textbf{Post-trained} & \textbf{SLR}\\
\hline

Yellowstone
&
Acadia National Park
\\
\hline

Yellowstone National Park
&
Yosemite
\\
\hline

Yellowstone
&
Everglades National Park
\\
\hline
Yellowstone
&
Rocky Mountain National Park
\\
\hline
Yellowstone National Park
&
Great Smoky Mountains National Park
\\
\hline
Yellowstone National Park
&
Glacier National Park
\\
\hline
Yellowstone National Park
&
Grand Teton National Park
\\
\hline
Yellowstone
&
Yellowstone National Park
\\
\hline\hline
\end{tabular}
\end{table*}

\begin{table*}[!htp]\centering
\caption{An example reasoning question where the post-trained model is unable to find the correct answer in 64 samples while SLR succeeds, using Llama3.1-8B.}
\label{tbl:qual_reasoning}
\footnotesize
\setlength{\tabcolsep}{10pt}
\renewcommand{\arraystretch}{1.15}

\begin{tabular}{p{0.45\textwidth} p{0.45\textwidth}}
\hline\hline
\multicolumn{2}{p{\dimexpr0.96\textwidth\relax}}{%
\textit{Prompt: ``A company pays each of its employees \$600 in a month. The company has a policy of increasing the salaries of each of its employees by 10\% of the initial salary every year for those who've stayed in the company for five years. If Sylvie just clocked 5 years in the company last December, what's her annual salary after three more years of service?''}%
}\\
\hline

\multicolumn{2}{p{\dimexpr0.96\textwidth\relax}}{%
\textit{In a month, Sylvie earns \$600, which totals \$600/month $\times$ 12 months/year $= \$7200$ in a year. Sylvie's earnings will increase by $\frac{10}{100} \times \$7200/\text{year} = \$720/\text{year}$ after 5 years. In three years, the total increase in her salary will be $3$ years $\times \$720/\text{year} = \$2160$. After three years, Sylvie will be paid $\$7200 + \$2160 = \$9360$.}%
}\\
\hline

\textbf{Post-trained} & \textbf{SLR}\\
\hline
To find Sylvie's annual salary after three more years of service, we need to calculate the salary after the 10\% increase, which happens every year for 5 years.

Her initial annual salary is \$600.

1. Increase after 1st year: \$600 + (10\% of \$600) = \$600 + \$60 = \$660

2. Increase after 2nd year: \$660 + (10\% of \$660) = \$660 + \$66 = \$726

3. Increase after 3rd year: \$726 + (10\% of \$726) = \$726 + \$72.6 = \$798.6

4. Increase after 4th year: \$798.6 + (10\% of \$798.6) = \$798.6 + \$79.86 = \$878.46

5. Increase after 5th year: \$878.46 + (10\% of \$878.46) = \$878.46 + \$87.846 = \$966.306

After the 5th year, Sylvie's annual salary becomes \$966.31.

For the next three years, she will still be getting a 10\% increase every year.

6. Increase after 6th year: \$966.31 + (10\% of \$966.31) = \$966.31 + \$96.631 = \$1062.941

7. Increase after 7th year: \$1062.94 + (10\% of \$1062.94) = \$1062.94 + \$106.294 = \$1169.234

8. Increase after 8th year: \$1169.23 + (10\% of \$1169.23) = \$1169.23 + \$116.923 = \$1286.153

After three more years of service, Sylvie's annual salary will be approximately \$1286.15.
&
To calculate Sylvie's salary after 3 years, we first need to find out her initial salary.

We are given that the company pays each employee \$600 in a month. Assuming there are 12 months in a year, the annual salary of Sylvie would be 600 * 12 = \$7{,}200.

Now, since Sylvie stayed in the company for 5 years, her salary would increase by 10\% of the initial salary every year for the next three years.

10\% of \$7{,}200 = (10/100) * 7{,}200 = \$720.

So, after the first year, Sylvie's salary would be \$7{,}200 + \$720 = \$7{,}920.

Now, to calculate her salary for the next two years, we will add \$720 to the previous year's salary.

After 2 years, her salary would be \$7{,}920 + \$720 = \$8{,}640.

After 3 years, her salary would be \$8{,}640 + \$720 = \$9{,}360.
\\
\hline\hline
\end{tabular}
\end{table*}

\end{document}